\documentclass[10pt,twocolumn,letterpaper]{article}

\usepackage{algorithm}
\usepackage{algorithmic}
\usepackage[utf8]{inputenc}

\usepackage{tabularx}
\usepackage{multirow}
\usepackage[export]{adjustbox}

\usepackage{pgfplots}
\usepackage{pgfplotstable}
\pgfplotsset{
    compat=1.18,  
    width=8cm,    
    height=6cm,   
    tick label style={font=\small},  
    label style={font=\small},       
    title style={font=\small\bfseries} 
}
\usepgfplotslibrary{statistics} 

\usepackage{graphicx}
\usepackage{subcaption}     
\usepackage{cvpr}              

\usepackage{soul}
\usepackage{xcolor}
\sethlcolor{yellow}

\definecolor{cvprblue}{rgb}{0.21,0.49,0.74}
\usepackage[pagebackref,breaklinks,colorlinks,allcolors=cvprblue]{hyperref}

\def\paperID{} 
\def\confName{CVPR}
\def\confYear{2026}

\title{
Exploring 6D Object Pose Estimation with Deformation
}

\author{%
	{Zhiqiang Liu $^1$, \quad Rui Song $^1$, \quad Duanmu Chuangqi $^{1}$, \quad Jiaojiao Li $^1$, \quad David Ferstl $^{2}$, \quad Yinlin Hu $^{2}$} \\
	{\normalsize $^1$ State Key Laboratory of ISN, Xidian University \quad \quad $^2$ MagicLeap} \\
}

\begin{document}

\twocolumn[{
\renewcommand\twocolumn[1][]{#1}
\maketitle
\vspace{-3mm}  
\begin{center}
    \setlength{\tabcolsep}{1pt}  
    \begin{tabular}{cccccccc}
        \multicolumn{8}{c}{\includegraphics[width=1.0\linewidth, keepaspectratio]{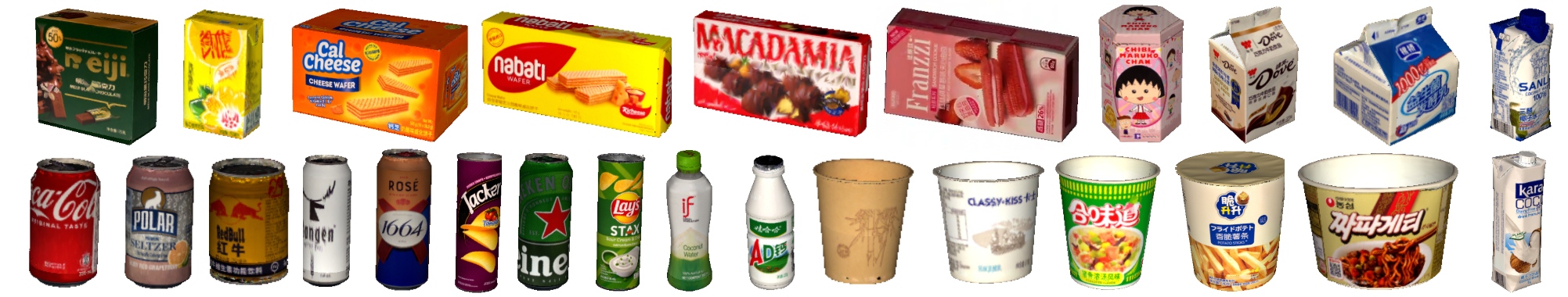}} \\

        \includegraphics[width=0.12\linewidth, height=0.15\linewidth]{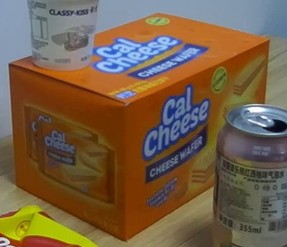} &
        \includegraphics[width=0.12\linewidth, height=0.15\linewidth]{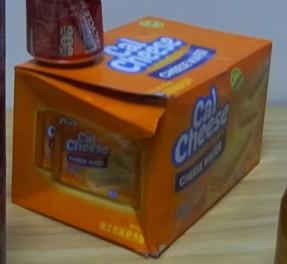} &
        \includegraphics[width=0.12\linewidth, height=0.15\linewidth]{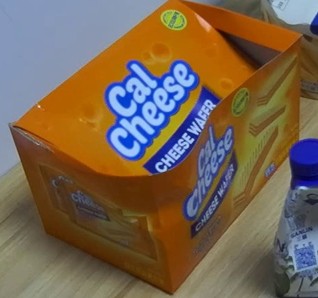} &
        \includegraphics[width=0.12\linewidth, height=0.15\linewidth]{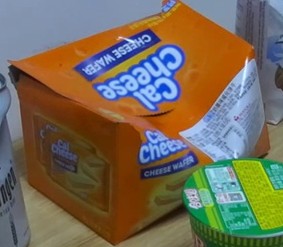} &
        \fbox{\includegraphics[width=0.12\linewidth, height=0.15\linewidth]{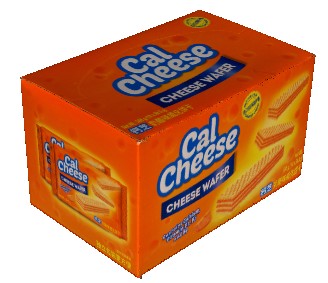}} &
        \fbox{\includegraphics[width=0.12\linewidth, height=0.15\linewidth]{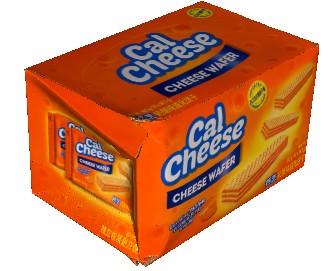}} &
        \fbox{\includegraphics[width=0.12\linewidth, height=0.15\linewidth]{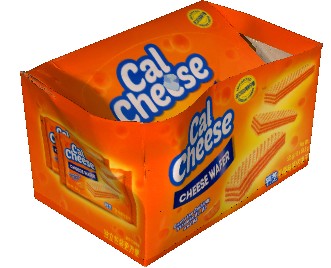}} &
        \fbox{\includegraphics[width=0.12\linewidth, height=0.15\linewidth]{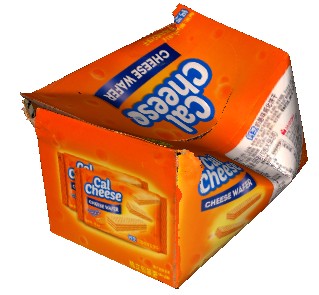}} \\
        
        \includegraphics[width=0.12\linewidth, height=0.16\linewidth]{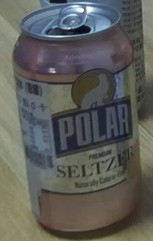} &
        \includegraphics[width=0.12\linewidth, height=0.16\linewidth]{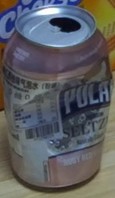} &
        \includegraphics[width=0.12\linewidth, height=0.16\linewidth]{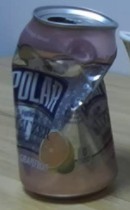} &
        \includegraphics[width=0.12\linewidth, height=0.16\linewidth]{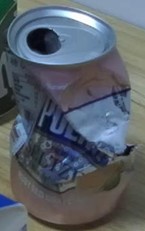} &
        \fbox{\includegraphics[width=0.12\linewidth, height=0.16\linewidth]{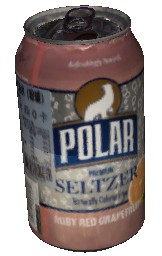}} &
        \fbox{\includegraphics[width=0.12\linewidth, height=0.16\linewidth]{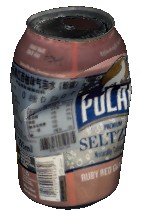}} &
        \fbox{\includegraphics[width=0.12\linewidth, height=0.16\linewidth]{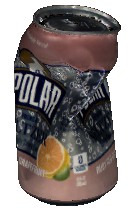}} &
        \fbox{\includegraphics[width=0.12\linewidth, height=0.16\linewidth]{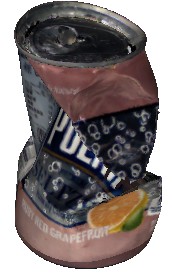}} \\

        {\small Canonical} & {\small Deformed 1} & {\small Deformed 2} & {\small Deformed 3} & {\small Canonical} & {\small Deformed 1} & {\small Deformed 2} & {\small Deformed 3} \\

    \end{tabular}
    \vspace{-1.5mm}
    \captionof{figure}{\textbf{6D object pose with deformation.} Object rigidity is a core assumption in 6D object pose estimation. While, many objects commonly regarded as rigid can undergo deformation over time due to factors such as collisions, wear from daily use, or improper handling during transport. In this work, we introduce a dataset specifically designed to capture such deformations for 6D object pose estimation. The dataset comprises scanned 3D meshes of 26 everyday objects, each represented in multiple deformed states, with precise mesh alignment across these states. Additionally, we provide 6D pose annotations for these objects in 133K frames, resulting in a total of 665K pose annotations, captured under a variety of conditions. We present the 26 canonical meshes (above), and some examples of captured images (left) and the corresponding scanned meshes (right) showing different levels of deformation.
    }
    \label{fig:teaser}
\end{center}
}]

\begin{abstract}
We present DeSOPE, a large-scale dataset for 6DoF deformed objects. Most 6D object pose methods assume rigid or articulated objects, an assumption that fails in practice as objects deviate from their canonical shapes due to wear, impact, or deformation. To model this, we introduce the DeSOPE dataset, which features high-fidelity 3D scans of 26 common object categories, each captured in one canonical state and three deformed configurations, with accurate 3D registration to the canonical mesh. Additionally, it features an RGB-D dataset with 133K frames across diverse scenarios and 665K pose annotations produced via a semi-automatic pipeline. We begin by annotating 2D masks for each instance, then compute initial poses using an object pose method, refine them through an object-level SLAM system, and finally perform manual verification to produce the final annotations. We evaluate several object pose methods and find that performance drops sharply with increasing deformation, suggesting that robust handling of such deformations is critical for practical applications. The project page and dataset are available at \href{https://desope-6d.github.io/}{https://desope-6d.github.io/}.
\end{abstract}    
\section{Introduction}
\label{sec:intro}

Estimating the 6DoF pose of objects is a core task in robotics~\cite{liang2025dynamicpose,li2025vita}, mixed reality~\cite{wu2025surgpose,jiang2025hand,pang2025splatpose}, and embodied AI~\cite{zhang2025adg,lee2025any6d,wang2025clip}. While existing benchmarks and methods~\cite{hodan2017tless, calli2015ycb, tyree20226hope, banerjee2025hot3d, hodan2018bop} have advanced the field significantly, they largely assume objects are perfectly rigid and match idealized CAD or scanned models—an assumption that rarely holds in practice. Models trained on perfect canonical meshes expect input images to align with these ideal shapes, yet real-world objects often deform unpredictably. In everyday settings, nominally ``rigid'' items like cardboard boxes, plastic bottles, and metal cans are frequently bent, dented, crushed, or partially collapsed due to regular use or rough handling, as shown in Fig.~\ref{fig:teaser}.

\begin{figure}[t]
    \centering
    \setlength\tabcolsep{1pt}
    \begin{tabular}{ccccccc}
        \includegraphics[width=0.235\linewidth, height=0.235\linewidth]{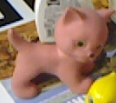} &
        \fbox{\includegraphics[width=0.235\linewidth, height=0.235\linewidth]{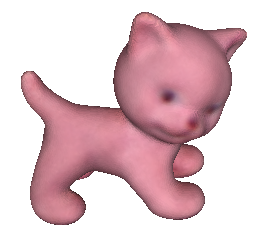}} &
        \includegraphics[width=0.235\linewidth, height=0.235\linewidth]{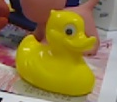} &
        \fbox{\includegraphics[width=0.235\linewidth, height=0.235\linewidth]{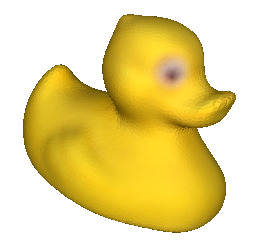}} & \\

        \includegraphics[width=0.235\linewidth, height=0.235\linewidth]{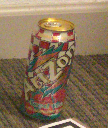} &
        \includegraphics[width=0.235\linewidth, height=0.235\linewidth]{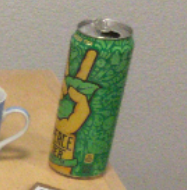} &
        \includegraphics[width=0.235\linewidth, height=0.235\linewidth]{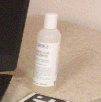} &
        \fbox{\includegraphics[width=0.235\linewidth, height=0.235\linewidth]{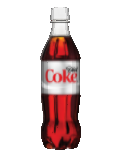}} & \\

        \includegraphics[width=0.235\linewidth, height=0.235\linewidth]{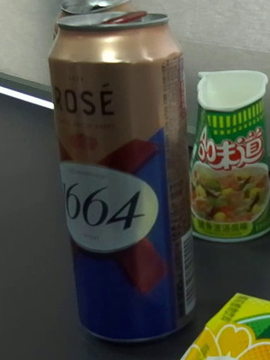} &
        \includegraphics[width=0.235\linewidth, height=0.235\linewidth]{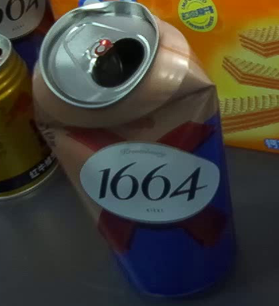} &
        \includegraphics[width=0.235\linewidth, height=0.235\linewidth]{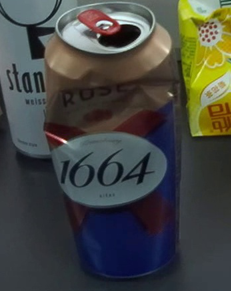} &
        \includegraphics[width=0.235\linewidth, height=0.235\linewidth]{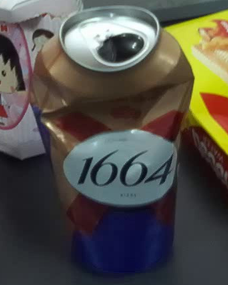} & \\

        \fbox{\includegraphics[width=0.235\linewidth, height=0.235\linewidth]{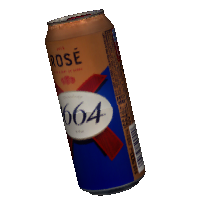}} &
        \fbox{\includegraphics[width=0.235\linewidth, height=0.235\linewidth]{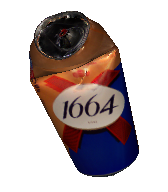}} &
        \fbox{\includegraphics[width=0.235\linewidth, height=0.235\linewidth]{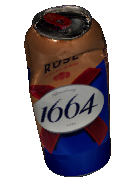}} &
        \fbox{\includegraphics[width=0.235\linewidth, height=0.235\linewidth]{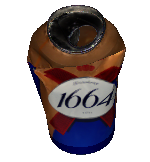}} & \\

    \end{tabular}
    \caption{{\bf Comparison of 6D object pose datasets.} The first row shows two examples from an instance-level dataset~\cite{brachmann2014linemod}, where each object instance is associated with its own 3D model, under the assumption that the object is perfectly rigid and does not deform over time. The second row depicts a category-level dataset~\cite{calli2015ycb}, in which multiple instances of the same category share a single 3D model (rightmost). The third row shows the canonical instance from the proposed DeSOPE dataset along with three deformed versions, while the fourth row presents their corresponding 3D meshes. We visualize meshes as black-bordered images on a white background in this figure.
    }
    \label{fig_dataset_compare}
\end{figure}

Current instance-level datasets and methods~\cite{maji2024yolopose, xu20246d, chen2022epro} offer only limited support for analyzing this regime. They largely focus on intact objects and interpret mesh variation as differences between separate rigid instances. 
On the other hand, although most category-level pose datasets and methods~\cite{lin2024instance, liu2025diff9d, cai2025gs, li2025gce,ren2025rethink} provide a canonical mesh to represent multiple instances within a category, they lack accurate meshes for individual instances and fail to capture instance-specific geometric variations, as shown in Fig.~\ref{fig_dataset_compare}.

We introduce DeSOPE, a real-world RGB-D dataset and 3D asset collection for deformed 6DoF object pose estimation, focusing on everyday items that are nominally rigid but often deformed. For 26 object categories, we capture high-quality scans of one canonical instance and three deformed variants (mild, moderate, severe), aligning all deformed meshes to their canonical counterparts using a flow-driven 3D registration framework~\cite{wang2025scflow2}. We then collect 133K RGB-D frames across different scenes with these objects. For 6D pose annotation, we label 2D instance masks~\cite{ravi2025sam}, generate initial poses with an object pose estimator~\cite{wen2024foundationpose}, and refine them by jointly optimizing object poses and an implicit neural shape representation~\cite{wang2023coslam}. After manual verification, DeSOPE provides 665K high-quality pose annotations across 104 deformed instances.

Using our dataset, we systematically evaluate several typical 6D object pose methods that assume input images align with perfect canonical meshes, which does not hold in this case. Our results show that performance drops significantly as objects deviate from their canonical shapes, highlighting deformation as a major, underexplored limitation in current 6D pose pipelines. To the best of our knowledge, DeSOPE is the first dataset to explicitly capture deformation for 6D object pose estimation.

\section{Related Work}
\label{sec:related_work}

\noindent {\bf Instance-level 6D object pose datasets}, including LINEMOD~\cite{brachmann2014linemod}, T-LESS~\cite{hodan2017tless}, YCB-V~\cite{calli2015ycb}, HOPE~\cite{tyree20226hope}, and others~\cite{fu2022wild6d, ahmadyan2021objectron, reizenstein2021co3d, iros23handal}, have been pivotal in advancing object pose estimation. In these datasets, each object instance is paired with its own 3D model—typically a clean CAD geometry or high-quality scan—and is assumed to be perfectly rigid. This one-to-one mapping facilitates precise pose supervision and standardized evaluation across methods.

However, these datasets mostly feature intact, undeformed objects in relatively controlled settings. Even with occlusions, clutter, or hand interactions, each object is assumed to perfectly match its reference model, leaving methods untested against real-world deviations. By contrast, DeSOPE provides carefully deformed versions of nominally rigid objects, explicitly aligned to their canonical counterparts, enabling evaluation under realistic deformation.

\noindent {\bf Category-level object pose datasets} aim to model the generalization beyond individual instances by grouping objects into semantic categories~\cite{wang2019nocs, jung2024housecat6d}. Typically, all instances within a category are represented by a single canonical mesh or template, creating a one-to-many mapping in which diverse objects are assumed to roughly align with the same 3D shape. These datasets are valuable for studying category-level evaluation and for enabling pose estimation when instance-specific CAD models are unavailable.

However, this design limits the study of deformation. Without accurate 3D meshes for each instance, we cannot capture how an object’s pose and geometry change when bent, dented, or compressed in real-world scenarios. By contrast, DeSOPE provides scanned 3D meshes of objects in multiple deformation states, along with precise registration between each deformed mesh and its canonical counterpart. Table~\ref{tab:objaverse_comparison} summarizes the comparison of different 6D object pose datasets.

\noindent {\bf Deformable and non-rigid targets} have been widely studied in domains such as garments~\cite{chen2025non,chen2025metafold,avigal2022speedfolding,canberk2023cloth}, soft bodies~\cite{liu2025spatial,chen2025dv,liu2024softmac,sorensen2023soft}, and articulated humans~\cite{yu2025dphuman, chen2025dv,bragagnolo2024multi,gombolay2024human}, where deformation is expected and serves as the primary focus. In contrast, our work targets objects that are nominally rigid—such as packaging and containers—but frequently appear deformed due to damage, wear, or everyday handling. These deviations challenge methods that assume input images match perfect canonical meshes, often causing performance to degrade when objects deform unpredictably in practice.
DeSOPE is designed to fill this gap, offering resources that enable the evaluation and future development of deformation-aware 6D object pose estimation techniques.

\begin{table}
  \centering
  \small 
  \setlength{\tabcolsep}{4pt} 
        \begin{tabular}{lcccc}
            \toprule
                    & Type          & Categories & Instances & Images \\
            \midrule
            LINEMOD~\cite{brachmann2014linemod}  & Instance      & -       & 15      & 19K    \\ 
            YCBV~\cite{calli2015ycb}         & Instance    & -       & 21      & 80K    \\
            PhoCaL~\cite{wang2022phocal}         & Instance     & 8       & 60      & 3K     \\
            Wild6D~\cite{fu2022wild6d}           & Instance     & 5       & 162     & 10K    \\
            Objectron~\cite{ahmadyan2021objectron} & Instance   & 9       & 17K     & 4M     \\
            CO3D~\cite{reizenstein2021co3d}      & Instance      & 5       & 19K     & 1.5M   \\
            HANDAL~\cite{iros23handal}           & Instance         & 17      & 212     & 308K   \\
            REAL275~\cite{wang2019nocs}          & Category     & 6       & 42      & 8K     \\
            HouseCat6D~\cite{jung2024housecat6d} & Category   & 10      & 194     & 24K    \\
            {\bf DeSOPE}                         & Deform.           & 26      & 104     & 133K   \\
            \bottomrule
        \end{tabular}
      \caption{{\bf Comparison of 6D object pose datasets.} Most instance-level datasets focus on individual instances, assuming objects are perfectly rigid and treating each instance independently. Category-level datasets, on the other hand, provide a single canonical mesh representing all instances within a semantic category. The proposed DeSOPE dataset captures multiple deformation states of the same instance, offering accurate 3D registration between the deformed state and the canonical form, along with precise 6D object pose annotations in images captured across diverse scenes.
    }
  \label{tab:objaverse_comparison}%
\end{table}%

\section{DeSOPE Dataset}
\label{sec:method}

\begin{figure*}[t]
  \centering
   \includegraphics[width=1.0\linewidth]{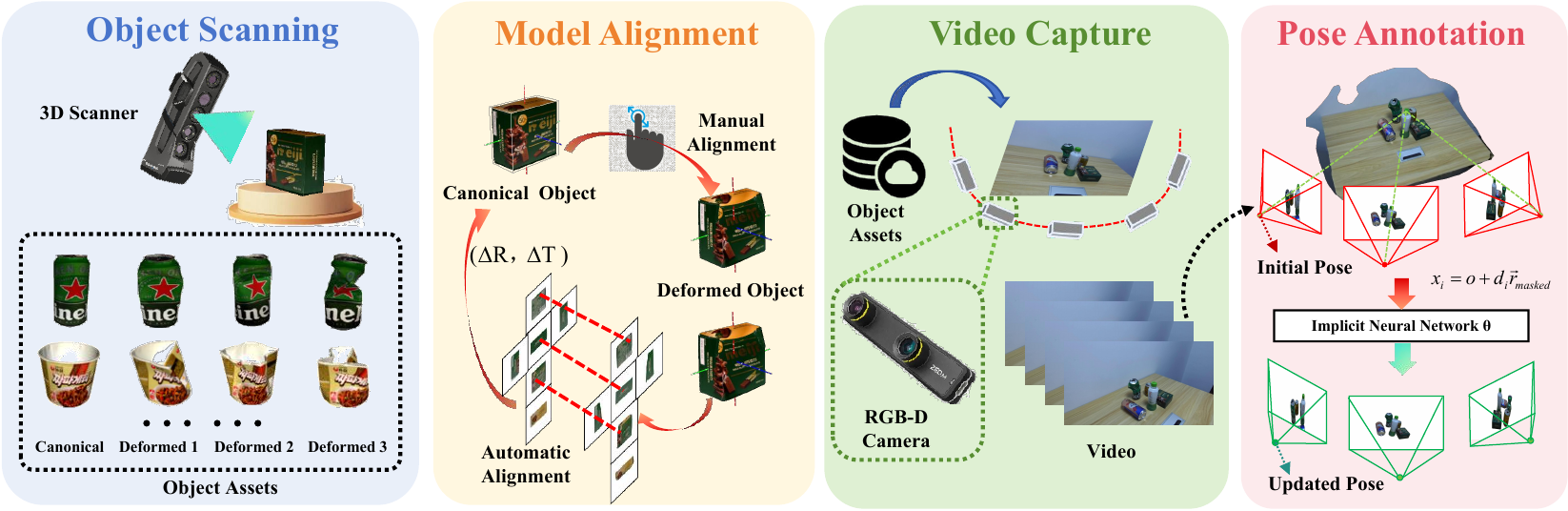}
   \caption{{\bf Overview of the dataset generation framework.} The framework consists of four main steps: {\bf Object Scanning}, which acquires the canonical mesh of objects along with multiple deformed states of the same instance using a high-precision 3D scanner; {\bf Model Alignment}, beginning with coarse manual alignment and followed by flow-driven 3D registration~\cite{yang2023scflow}; {\bf Video Capture}, which records RGB-D videos of objects across diverse scenes with a stereo camera; and {\bf Pose Annotation}, which performs initial object labeling and iteratively refines poses using implicit neural networks to obtain accurate annotations.
   }
   \label{fig_DeSOPE}
\end{figure*}

This section presents DeSOPE, a real-world dataset for 6D object pose estimation covering diverse object categories and deformable objects. We begin with the 3D scanning and registration process between canonical and deformed meshes in Section~\ref{3D_col_align}, and then describe the image acquisition and semi-automatic annotation procedures in Section~\ref{acq_annote}. Figure~\ref{fig_DeSOPE} illustrates the overall data collection and annotation framework.

\subsection{3D Model Scanning and Alignment}
\label{3D_col_align}
We select 26 daily object categories. Each category contains one canonical (undeformed) reference instance and three additional instances with progressively increasing deformation levels, categorized as mild, moderate, and severe, resulting in a total of 104 objects, as illustrated in Fig.~\ref{fig_DeSOPE}. 

We acquire the meshes of all instances using a high-precision Go!SCAN SPARK scanner~\cite{goscan_spark}, with an average scanning time of approximately 10 minutes per object.

After acquiring all 3D models, we first obtain an initial registration between the canonical mesh and each deformed mesh through manual alignment.
We then refine the alignment using a flow-guided matching strategy. Specifically, we render each mesh from six orthogonal viewpoints (front, back, left, right, top, and bottom), applying the same rotation and translation to ensure consistency across views. To align the deformed mesh with the canonical mesh, we compute dense 2D correspondences between pairs of rendered images from the same viewpoint.

We use SCFlow~\cite{yang2023scflow} to predict dense 2D correspondences. The model is pretrained on approximately 9 million rendered image pairs generated from 90K meshes, including ShapeNet-Objects~\cite{shapenet2015}, Google Scanned Objects~\cite{googleObjects2022}, and Objaverse~\cite{objaverse2023}.
SCFlow jointly estimates dense 2D correspondences and object pose in an iterative manner; however, in our experiments, we utilize only the 2D correspondence predictions during inference.

Finally, we lift the 2D correspondences to 3D by leveraging the inherent 2D–3D mappings established during rendering, thereby establishing 3D–3D correspondences between the canonical and deformed meshes. For registration, we adopt a two-step strategy: we first apply RANSAC~\cite{raguram2012ransac} to remove outliers and estimate initial transformation parameters, and then refine the transformation using the Umeyama~\cite{umeyama2002least} algorithm, which computes the optimal similarity transformation (including rotation, translation, and scale) over the inlier set, resulting in precise final mesh alignment.

\subsection{Video Capture and Pose Annotation}
\label{acq_annote}

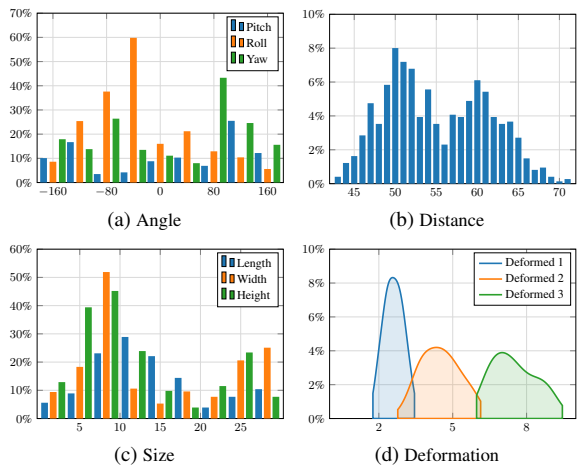
\begin{figure}[t]
    \centering
    \definecolor{cvprblue}{RGB}{31,119,180}
    \definecolor{cvprorange}{RGB}{255,127,14}
    \definecolor{cvprgreen}{RGB}{44,160,44}

    \begin{subfigure}[t]{0.45\linewidth}
        \centering
        \begin{tikzpicture}[scale=0.51]
            \begin{axis}[
                ybar,
                bar width=5pt,
                tick align=inside,
                xtick={-160,-80,0,80,160},
                ytick={0,10,20,30,40,50,60,70},
                xticklabel style={align=center, font=\small},
                xmin=-180, xmax=180,
                enlarge x limits={rel=0.01},
                ymin=0, ymax=70,
                yticklabel={\pgfmathprintnumber{\tick}\%},
                yticklabel style={font=\small},
                axis background/.style={fill=white},
                grid=major,
                grid style={gray!30},
                legend columns=1,
                legend cell align=left,
                legend style={font=\small, at={(0.98,0.98)}, anchor=north east},
            ]
            \addplot[fill=cvprblue, opacity=1, draw=none]
                table[x expr=\thisrow{BinCenter}+0.1, y=Pitch, col sep=comma] {rotation_histogram_9bins_v3.csv};
            \addlegendentry{Pitch}
            \addplot[fill=cvprorange, opacity=1, draw=none]
                table[x expr=\thisrow{BinCenter}, y=Roll, col sep=comma] {rotation_histogram_9bins_v3.csv};
            \addlegendentry{Roll}
            \addplot[fill=cvprgreen, opacity=1, draw=none]
                table[x expr=\thisrow{BinCenter}-0.1, y=Yaw, col sep=comma] {rotation_histogram_9bins_v3.csv};
            \addlegendentry{Yaw}
            \end{axis}
        \end{tikzpicture}
        \caption{\scriptsize Angle}
    \end{subfigure}%
    \hspace{0.05cm}
    \begin{subfigure}[t]{0.45\linewidth}
        \centering
        \begin{tikzpicture}[scale=0.51]
            \begin{axis}[
                ybar,
                bar width=0.16cm,
                xmin=42, xmax=72,
                tick align=inside,
                xtick={40,45,50,55,60,65,70},
                xticklabel style={align=center, font=\small},
                ymin=0, ymax=10,
                yticklabel={\pgfmathprintnumber{\tick}\%},
                yticklabel style={font=\small},
                axis background/.style={fill=white},
                grid=major,
                grid style={gray!30},
            ]
            \addplot[fill=cvprblue, opacity=1, draw=none]
                table[x index=4,y index=5,col sep=comma] {Rtable.csv};
            \end{axis}
        \end{tikzpicture}
        \caption{\scriptsize Distance}
    \end{subfigure}

    \vspace{2pt}

    \begin{subfigure}[t]{0.45\linewidth}
        \centering
        \begin{tikzpicture}[scale=0.51]
            \begin{axis}[
                ybar,
                bar width=5pt,
                tick align=inside,
                xtick={5,10,15,20,25},
                xticklabel style={align=center, font=\small},
                enlarge x limits={rel=0.01},
                xmin=0, xmax=30,
                ymin=0, ymax=60,
                ytick={0,10,20,30,40,50,60},
                yticklabel={\pgfmathprintnumber{\tick}\%},
                yticklabel style={font=\small},
                xlabel style={font=\small},
                axis background/.style={fill=white},
                grid=major,
                grid style={gray!30},
                legend columns=1,
                legend cell align=left,
                legend style={font=\small, at={(0.98,0.98)}, anchor=north east},
            ]
            \addplot[fill=cvprblue, opacity=1, draw=none]
                table[x expr=\thisrow{BinCenter}+0.1, y=Length, col sep=comma] {size_histogram_9bins_v2.csv};
            \addlegendentry{Length}
            \addplot[fill=cvprorange, opacity=1, draw=none]
                table[x expr=\thisrow{BinCenter}, y=Width, col sep=comma] {size_histogram_9bins_v2.csv};
            \addlegendentry{Width}
            \addplot[fill=cvprgreen, opacity=1, draw=none]
                table[x expr=\thisrow{BinCenter}-0.1, y=Height, col sep=comma] {size_histogram_9bins_v2.csv};
            \addlegendentry{Height}
            \end{axis}
        \end{tikzpicture}
        \caption{\scriptsize Size}
    \end{subfigure}%
    \hspace{0.05cm}
    \begin{subfigure}[t]{0.45\linewidth}
    \centering
        \begin{tikzpicture}[scale=0.51]
            \begin{axis}[
                xmin=0, xmax=10,
                xtick={2,5,8},
                ymin=0, ymax=10,
                xticklabel style={font=\small},
                yticklabel={\pgfmathprintnumber{\tick}\%},
                yticklabel style={font=\small},
                axis background/.style={fill=white},
                grid=major,
                grid style={gray!30},
                legend columns=1,
                legend cell align=left,
                legend style={font=\small, at={(0.98,0.98)}, anchor=north east},
            ]
            \addplot[cvprblue, solid, line width=1.2pt, mark=none, fill=cvprblue, fill opacity=0.15]
                table[x=x, y=deformation1, col sep=comma] {deformation1.csv} \closedcycle;
            \addlegendentry{Deformed 1}
            \addplot[cvprorange, solid, line width=1.2pt, mark=none, fill=cvprorange, fill opacity=0.15]
                table[x=x, y=deformation2, col sep=comma] {deformation2.csv} \closedcycle;
            \addlegendentry{Deformed 2}
            \addplot[cvprgreen, solid, line width=1.2pt, mark=none, fill=cvprgreen, fill opacity=0.15]
                table[x=x, y=deformation3, col sep=comma] {deformation3.csv} \closedcycle;
            \addlegendentry{Deformed 3}
            \end{axis}
        \end{tikzpicture}
        \caption{\scriptsize Deformation}
    \end{subfigure}
    \caption{
    {\bf Statistical analysis of the DeSOPE dataset.} We summarize the different properties of our dataset using histograms and report percentages (\%) on the y-axis across all subplots. {\bf (a)} Distribution of camera pose angles (x-axis: rotation angle in degrees), illustrating the coverage of pitch, roll, and yaw across all annotated frames. {\bf (b)} Distribution of object-to-camera distances (x-axis: distance in cm), with values concentrated around 50-60 cm. {\bf (c)} Distribution of physical dimensions for 104 object instances across 26 categories (x-axis: size in cm), demonstrating the diversity in object sizes. {\bf (d)} Distribution of deformation severity (x-axis: deformation magnitude in cm) across three levels: mild for Deformed 1, moderate for Deformed 2, and severe for Deformed 3. The deformation magnitude is computed as the average point-wise 3D distance between corresponding vertices after aligning the deformed and undeformed meshes.
    }
    \label{fig_statistic}
\end{figure}

\noindent \textbf{26 object categories.}
We collect 26 categories of daily objects, each comprising four instances: one canonical mesh and three deformed variants with increasing levels of deformation, resulting in a total of 104 objects, as illustrated in Fig.~\ref{fig:teaser}. The dataset covers a wide range of materials and geometric properties, including both regular and irregular shapes, as well as significant scale variation. We incorporate realistic deformation types—such as stretching, bending, compression, and twisting—and categorize them into three levels: mild, moderate, and severe. Fig.~\ref{fig_statistic} summarizes the statistical properties of the dataset.

\begin{figure*}[t]
    \centering
    \setlength\tabcolsep{1pt}
    \begin{tabular}{ccccccc}
        \includegraphics[width=0.16\linewidth, height=0.1\linewidth]{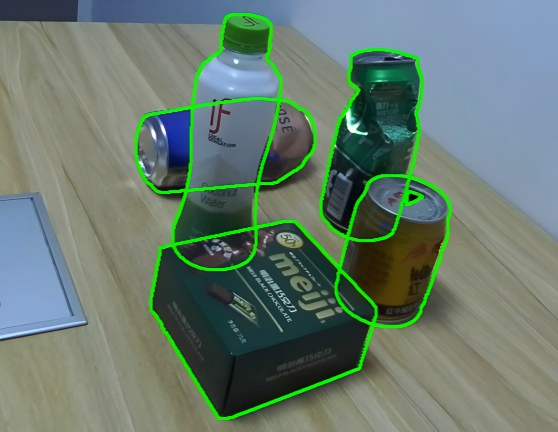} &
        \includegraphics[width=0.16\linewidth, height=0.1\linewidth]{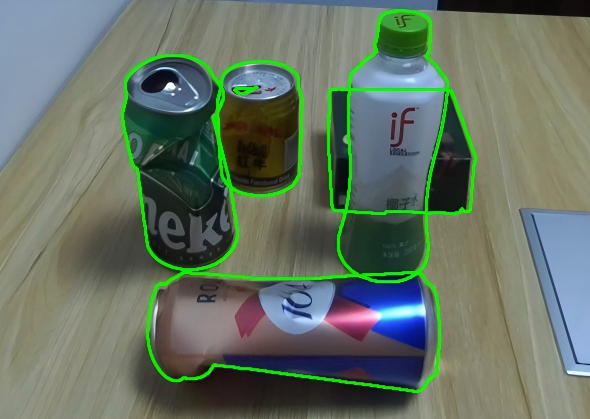} &
        \includegraphics[width=0.16\linewidth, height=0.1\linewidth]{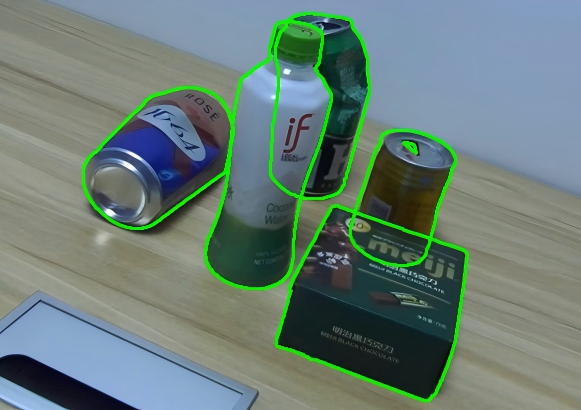} &
        \includegraphics[width=0.16\linewidth, height=0.1\linewidth]{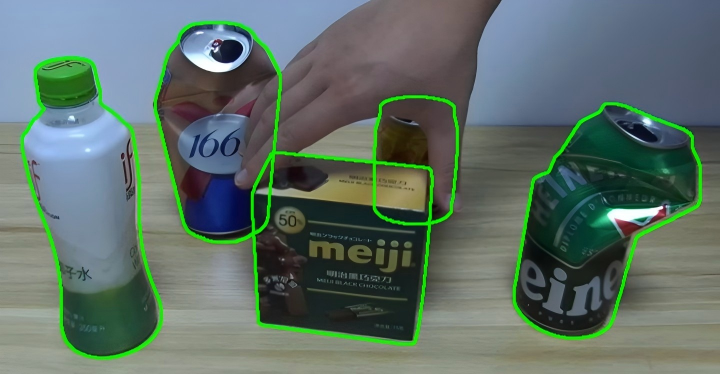} &
        \includegraphics[width=0.16\linewidth, height=0.1\linewidth]{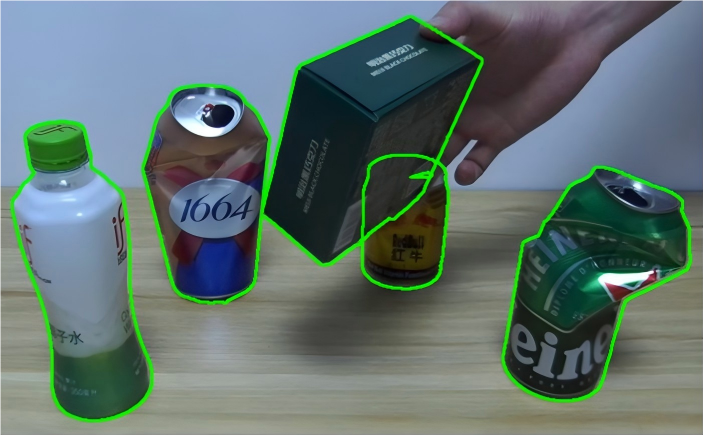} &
        \includegraphics[width=0.16\linewidth, height=0.1\linewidth]{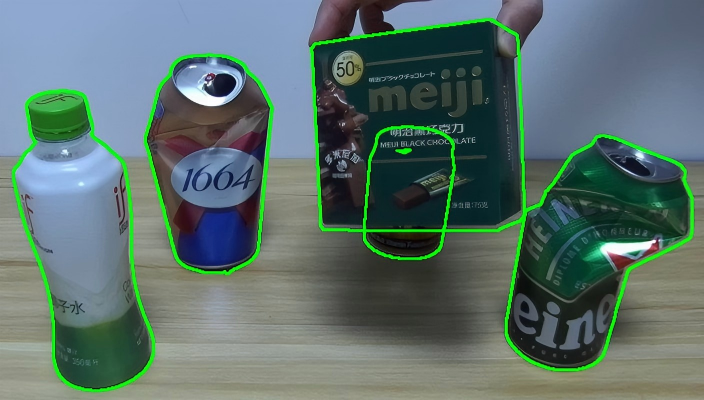} & \\

        \includegraphics[width=0.16\linewidth, height=0.1\linewidth]{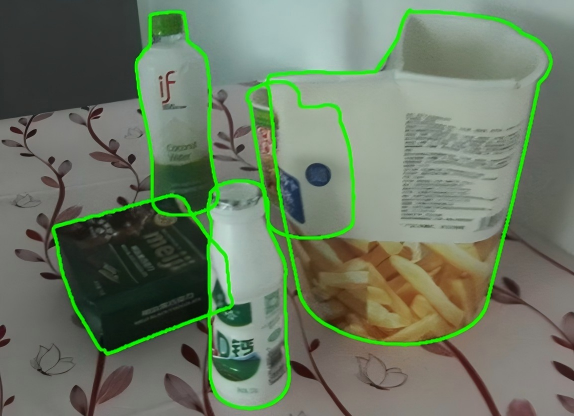} &
        \includegraphics[width=0.16\linewidth, height=0.1\linewidth]{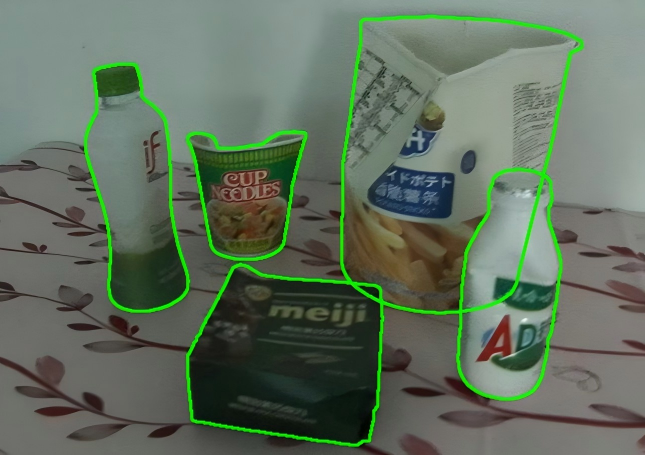} &
        \includegraphics[width=0.16\linewidth, height=0.1\linewidth]{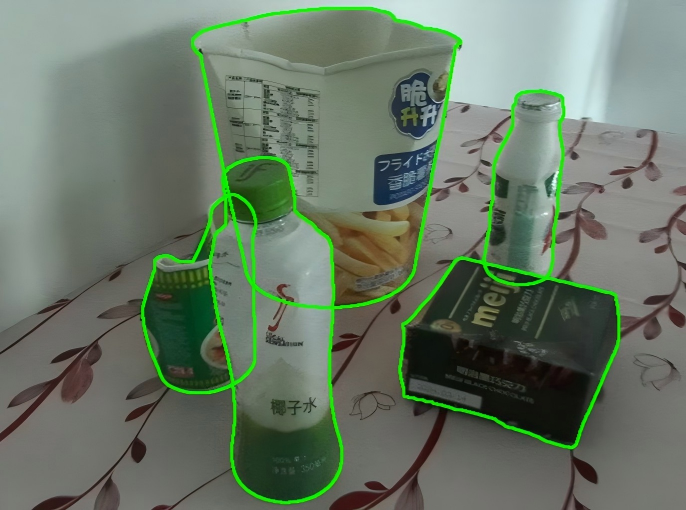} &
        \includegraphics[width=0.16\linewidth, height=0.1\linewidth]{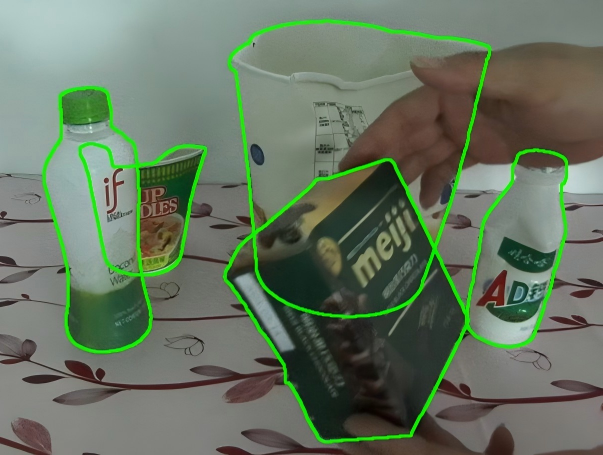} &
        \includegraphics[width=0.16\linewidth, height=0.1\linewidth]{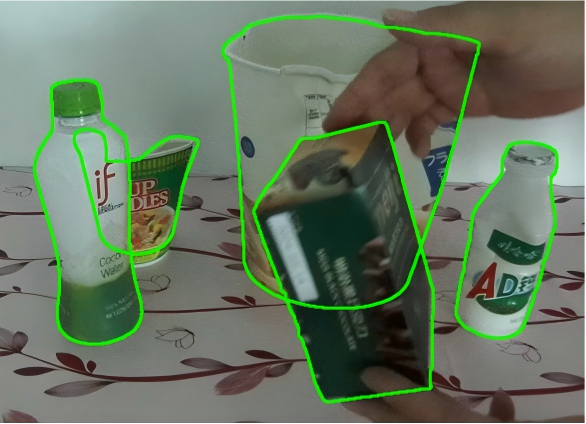} &
        \includegraphics[width=0.16\linewidth, height=0.1\linewidth]{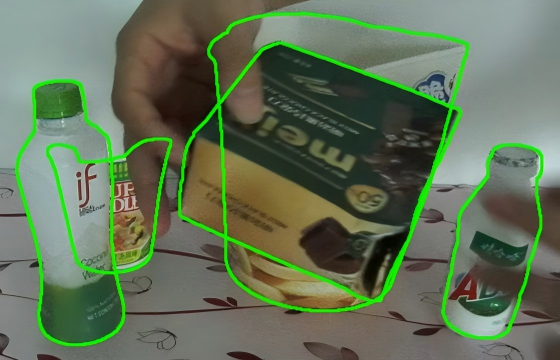} & \\

        \includegraphics[width=0.16\linewidth, height=0.1\linewidth]{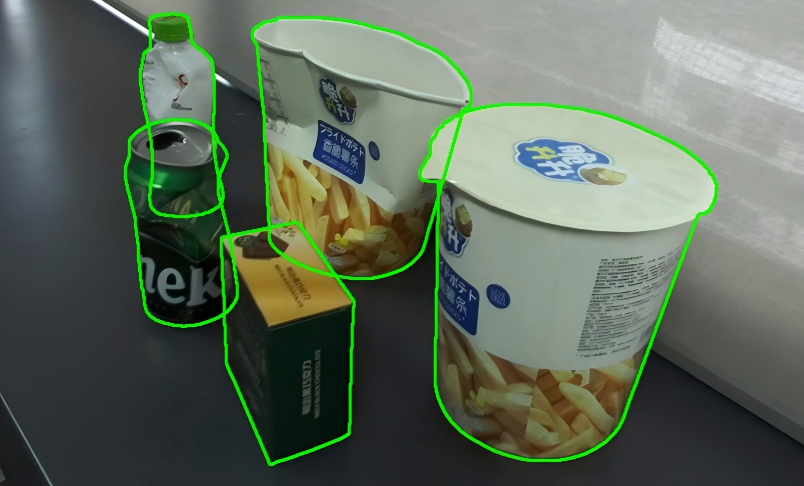} &
        \includegraphics[width=0.16\linewidth, height=0.1\linewidth]{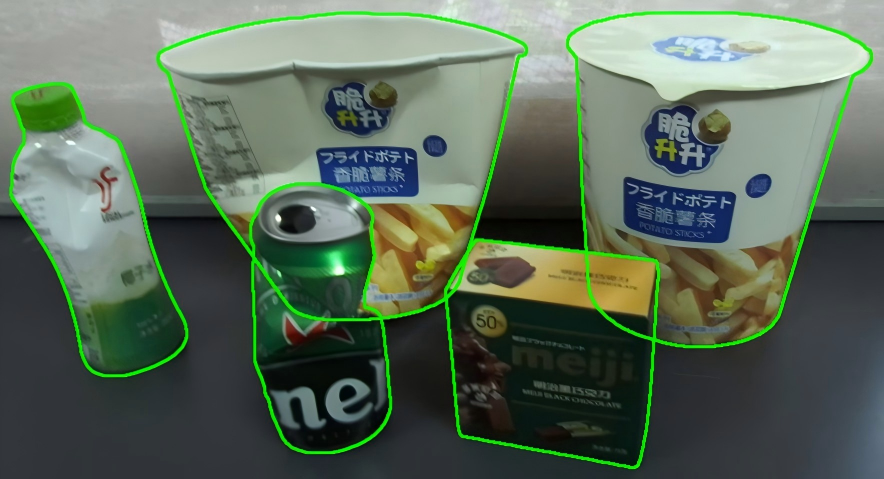} &
        \includegraphics[width=0.16\linewidth, height=0.1\linewidth]{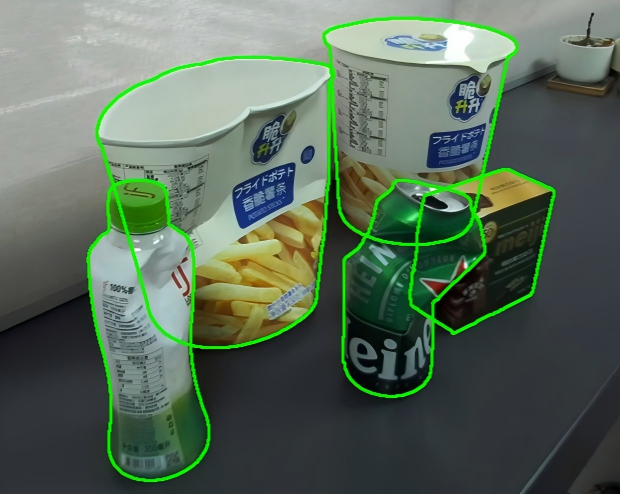} &
        \includegraphics[width=0.16\linewidth, height=0.1\linewidth]{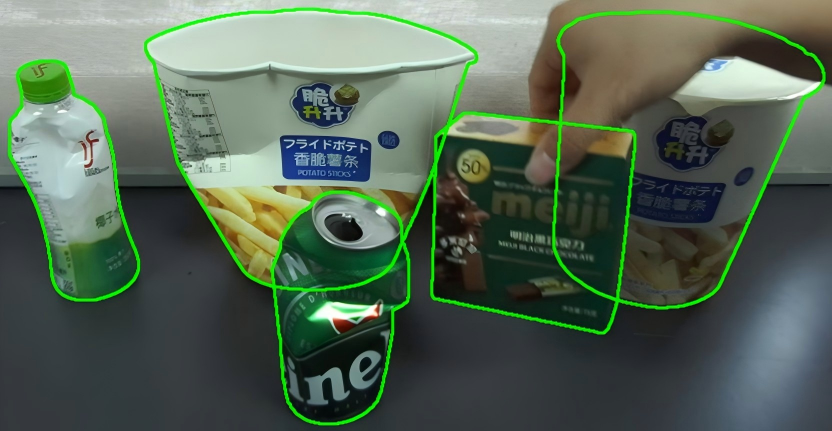} &
        \includegraphics[width=0.16\linewidth, height=0.1\linewidth]{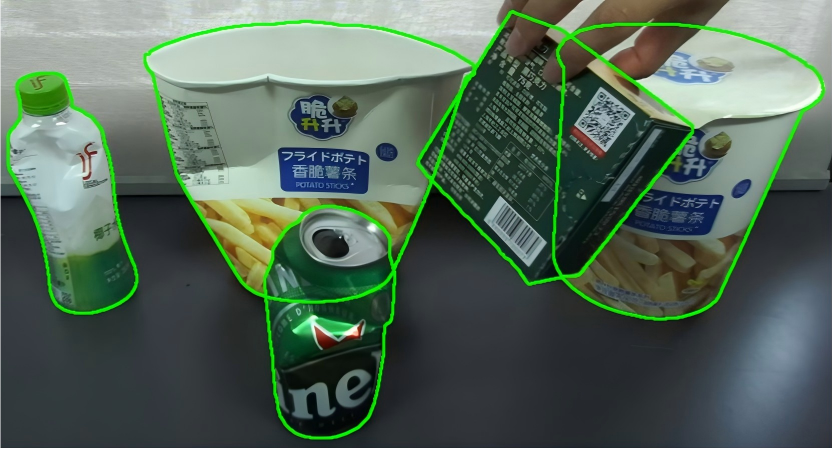} &
        \includegraphics[width=0.16\linewidth, height=0.1\linewidth]{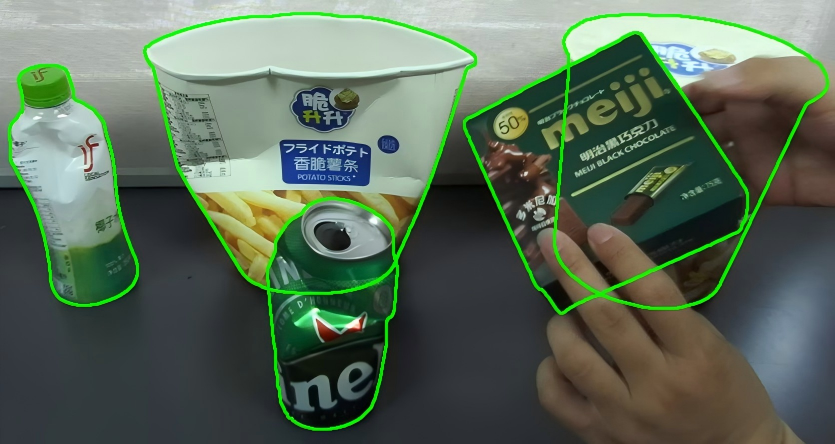} & \\

    \end{tabular}
    \caption{
    {\bf Example of captured images and pose annotations.} The green boundary contours represent pose projections onto the 2D plane using the corresponding mesh, as obtained by the annotation algorithm proposed in this paper. The dataset contains images captured in cluttered scenes under both human-manipulated and non-manipulated conditions.
    }
    \label{fig_annotation_results}
\end{figure*}

\noindent \textbf{Five daily scenarios.}
We collect the image dataset across multiple indoor environments, including a conference room, dining area, window-side setting, sofa area, and break room. 
For each scene, we record 208 videos: 104 in a static setting with the camera circling around the scene, and 104 in the same environment but with human manipulation, including actions such as picking up objects, holding, moving, shaking, and deliberately introducing hand-induced occlusions.
Example images and corresponding pose annotations are shown in Fig.~\ref{fig_annotation_results}.

\noindent \textbf{260 minutes of recordings.}
We capture the dataset using a ZED2i stereo camera~\cite{zed2i}. Each video is recorded for approximately 30 seconds at 30 FPS, yielding 120–240 sampled frames. In total, the dataset contains 133K valid RGB-D frames at a resolution of 1920×1080. Each frame contains about five object instances, including both deformed and undeformed objects, arranged in random combinations with inter-object occlusions. In total, the dataset provides 665K valid pose annotations.

\noindent\textbf{2D annotation and pose initialization.}
We begin the annotation process with 2D mask annotation.
Given an input RGB-D frame $(I_t, D_t)$ with camera intrinsics $K \in \mathbb{R}^{3 \times 3}$, we first employ a pre-trained segmentation model SAM2~\cite{ravi2025sam} to extract binary masks $\mathcal{M} = \{M_1, \ldots, M_n\}$ for all the $n$ instances in the scene, where $M_i \in \{0,1\}^{H \times W}$ indicates the $i$-th instance region. Unlike prior SLAM-based approaches~\cite{wang2023coslam} that rely on constant-speed motion models—which struggle under rapid motion and occlusion—we leverage FoundationPose~\cite{wen2024foundationpose} for robust initial pose estimation. FoundationPose estimates the relative pose of each instance via 2D-3D feature matching and PnP solvers, yielding candidate poses $\{\xi_{t,i}^{\text{init}}\}_{i=1}^5$. We perform consistency voting to eliminate outliers, retaining poses with pairwise errors below threshold $\tau$ ($5^\circ$ rotation , $5 cm$ translation) and computing the averaged initial pose.


\begin{figure*}[t]
    \centering
    \setlength\tabcolsep{1pt}
    \begin{tabular}{ccccccc}
        \fbox{\includegraphics[width=0.15\linewidth, height=0.15\linewidth]{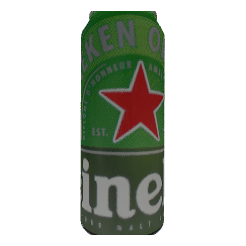}} &
        \fbox{\includegraphics[width=0.15\linewidth, height=0.15\linewidth]{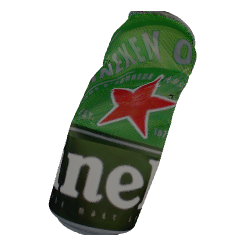}} &
        \fbox{\includegraphics[width=0.15\linewidth, height=0.15\linewidth]{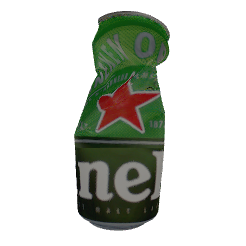}} &
        \fbox{\includegraphics[width=0.15\linewidth, height=0.15\linewidth]{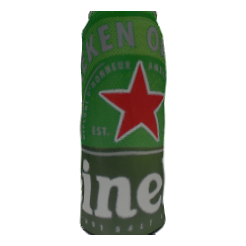}} &
        \includegraphics[width=0.15\linewidth, height=0.15\linewidth]{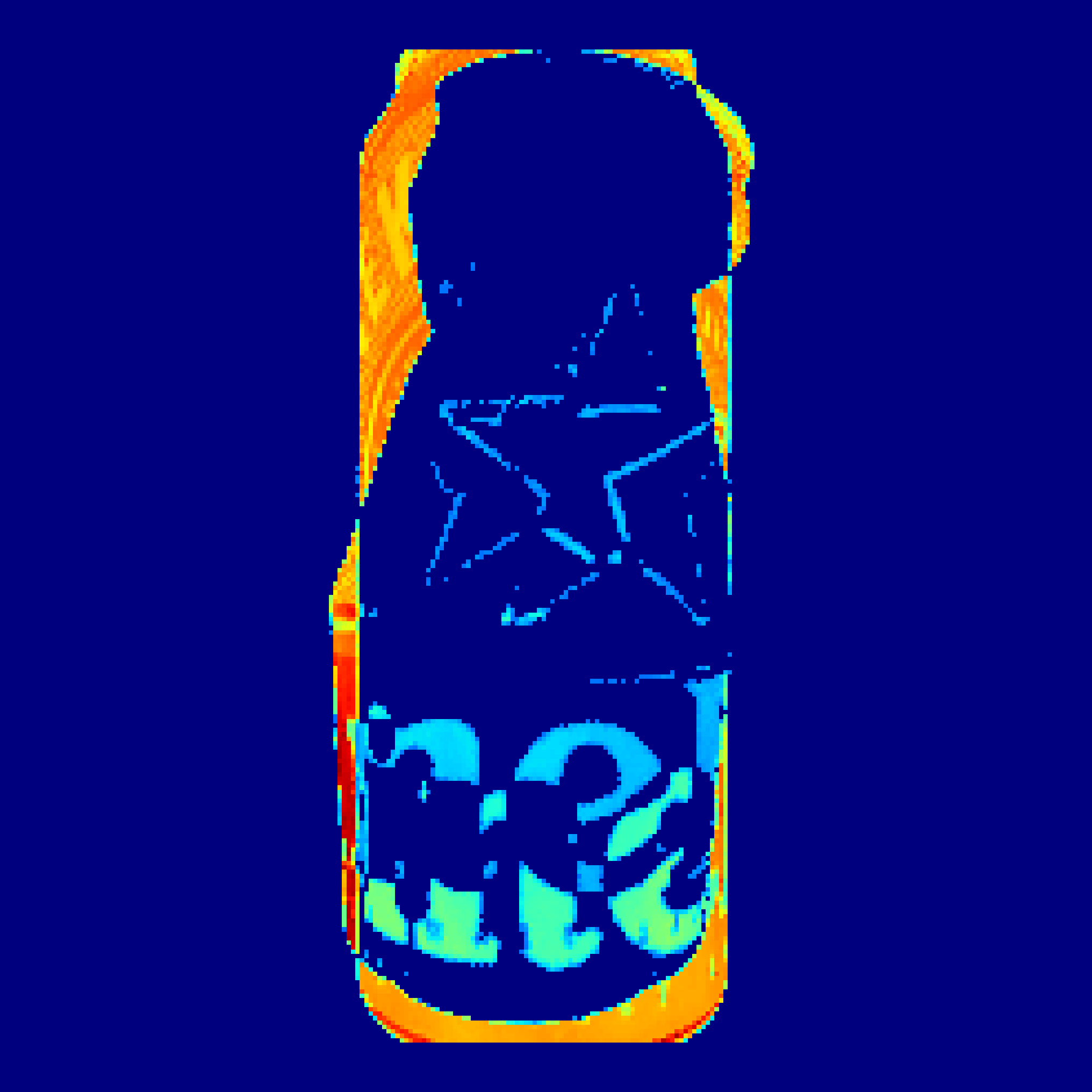} &
        \includegraphics[width=0.15\linewidth, height=0.15\linewidth]{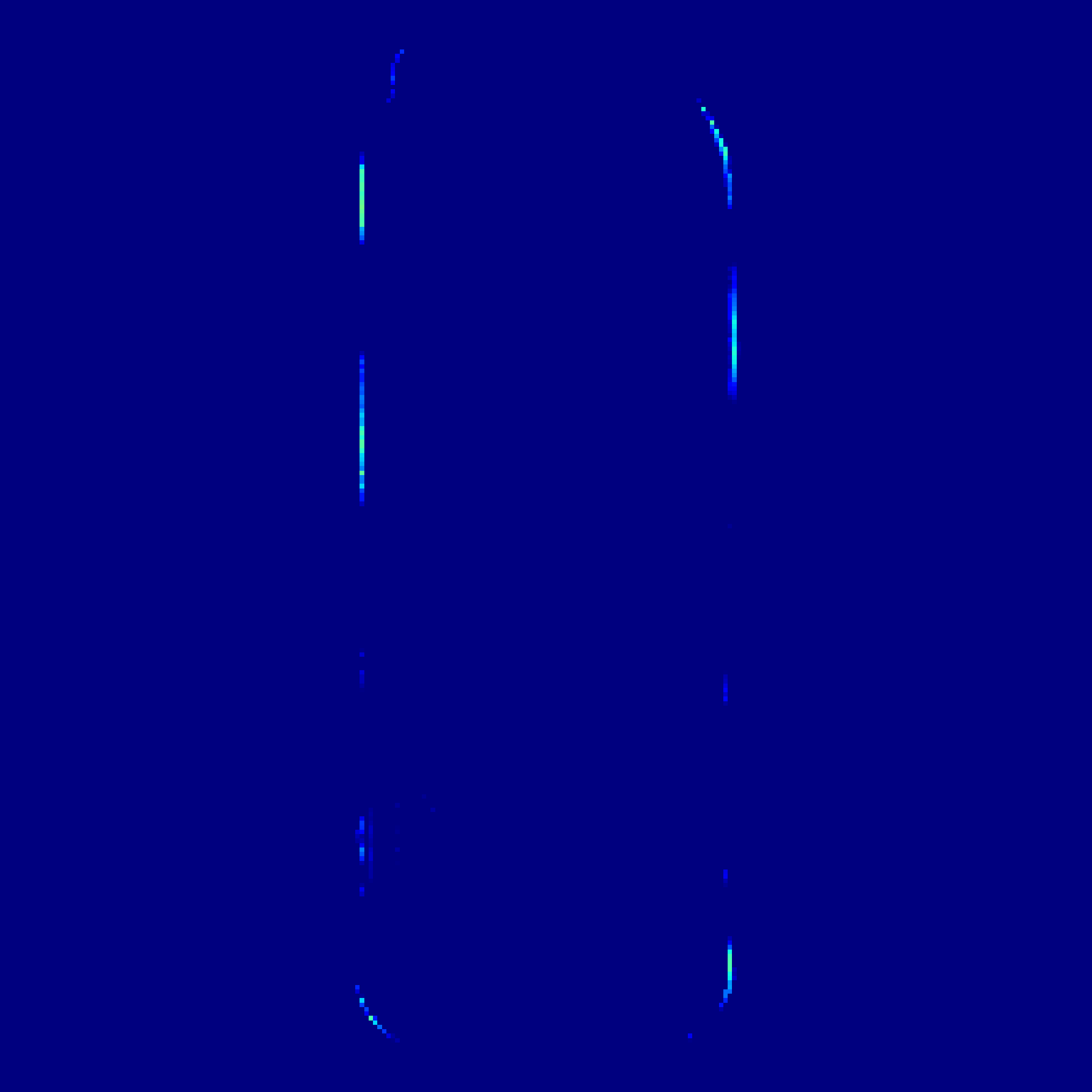} &
        \multirow{2}[0]{*}[0.13\linewidth]{\centering \includegraphics[width=0.05\linewidth, height=0.31\linewidth]{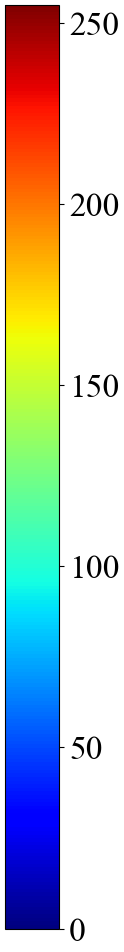}}\\

        \fbox{\includegraphics[width=0.15\linewidth, height=0.15\linewidth]{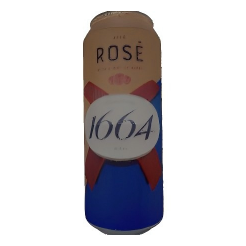}} &
        \fbox{\includegraphics[width=0.15\linewidth, height=0.15\linewidth]{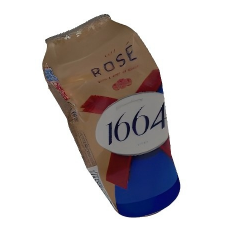}} &
        \fbox{\includegraphics[width=0.15\linewidth, height=0.15\linewidth]{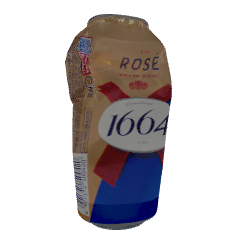}} &
        \fbox{\includegraphics[width=0.15\linewidth, height=0.15\linewidth]{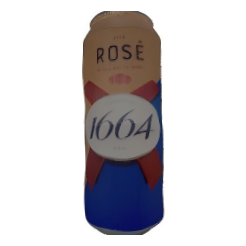}} &
        \includegraphics[width=0.15\linewidth, height=0.15\linewidth]{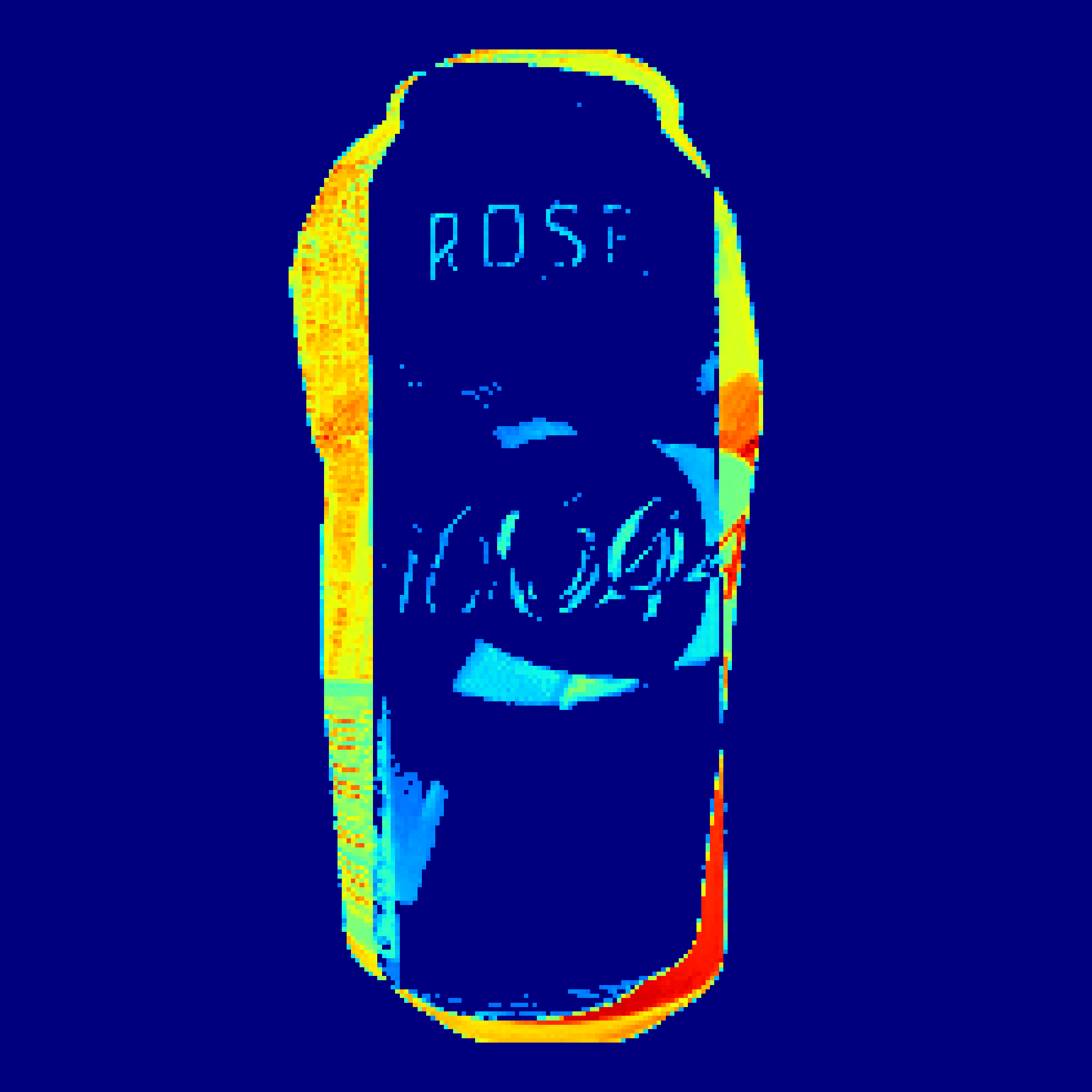} &
        \includegraphics[width=0.15\linewidth, height=0.15\linewidth]{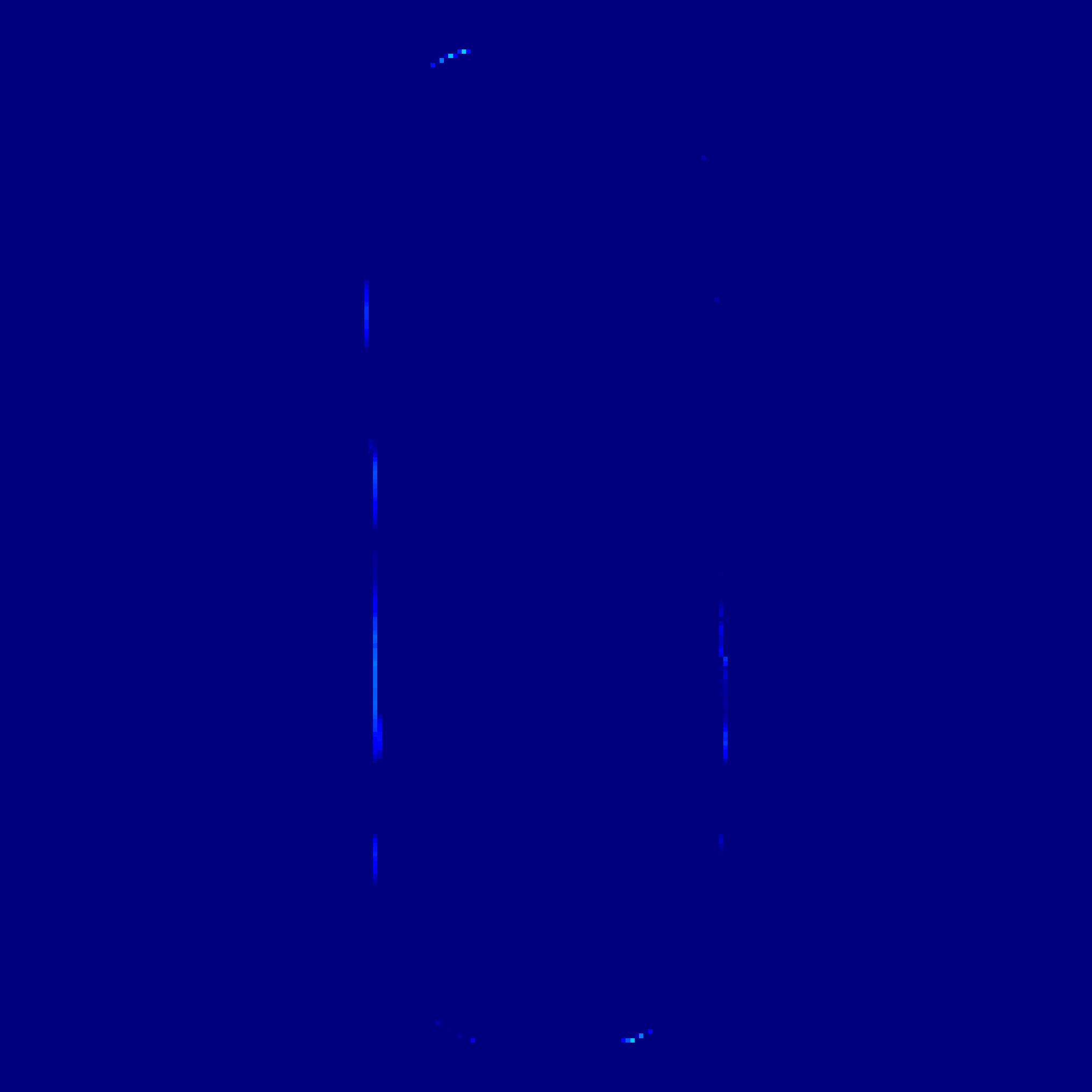} &
        \\  

        {\small Canonical mesh} &
        {\small Deformed mesh} &
        {\small +Manual} &
        {\small +Auto Refine} &
        {\small Manual error map} &
        {\small Refined error map} &
        {\small }  
    \end{tabular}
    \caption{{\bf Example of 3D model alignment.} We estimate the optimal registration between each deformed mesh and its corresponding canonical mesh through a two-stage process. We begin with a manual alignment to obtain a coarse initialization, providing a reasonable starting point for further refinement. This initial estimate is then refined based on dense 2D matching computed from six orthogonal viewpoints. The resulting error map visualizes the pixel-wise differences between the canonical mesh and the aligned deformed mesh when projected from the same viewpoints, both before and after refinement.
    }
    \label{fig_align_results}
\end{figure*}

\begin{figure*}[t]
    \centering
    \setlength\tabcolsep{1pt}
    \begin{tabular}{ccccccc}
        \includegraphics[width=0.16\linewidth, height=0.18\linewidth]{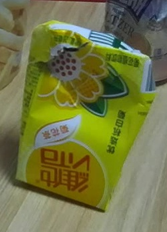} &
        \includegraphics[width=0.16\linewidth, height=0.18\linewidth]{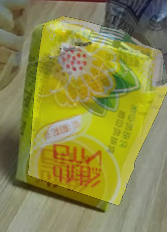} &
        \includegraphics[width=0.16\linewidth, height=0.18\linewidth]{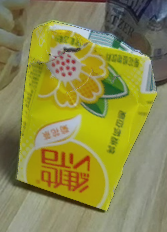} &
        \includegraphics[width=0.16\linewidth, height=0.18\linewidth]{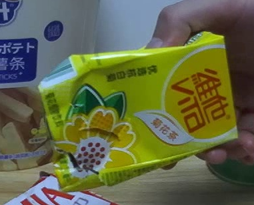} &
        \includegraphics[width=0.16\linewidth, height=0.18\linewidth]{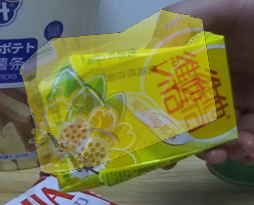} &
        \includegraphics[width=0.16\linewidth, height=0.18\linewidth]{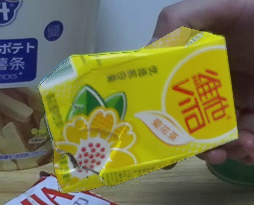} & \\

        \includegraphics[width=0.16\linewidth, height=0.18\linewidth]{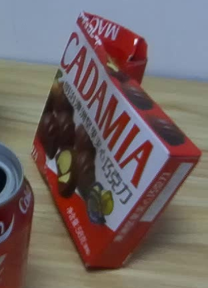} &
        \includegraphics[width=0.16\linewidth, height=0.18\linewidth]{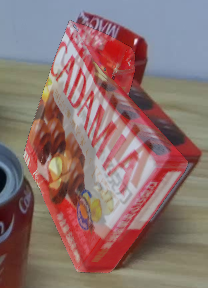} &
        \includegraphics[width=0.16\linewidth, height=0.18\linewidth]{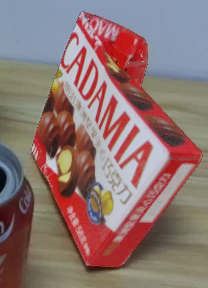} &
        \includegraphics[width=0.16\linewidth, height=0.18\linewidth]{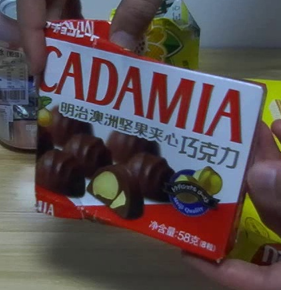} &
        \includegraphics[width=0.16\linewidth, height=0.18\linewidth]{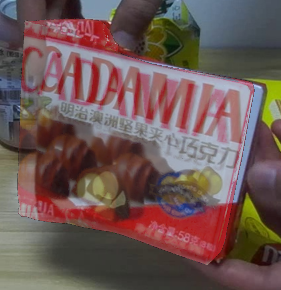} &
        \includegraphics[width=0.16\linewidth, height=0.18\linewidth]{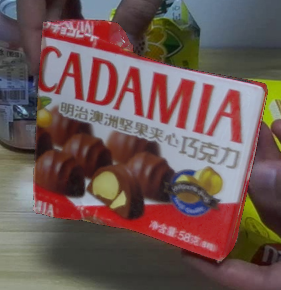} & \\

        {\small Input} & {\small Pose initialization} & {\small \bf Result} & {\small Input} & {\small Pose initialization} & {\small \bf Result}

    \end{tabular}
    \caption{
    {\bf Effect of pose refinement.} We visualize the predicted pose by overlaying the rendered textured mesh onto the input image according to the estimated object pose. The initial pose exhibits noticeable misalignment; after applying our pose refinement strategy, the rendered mesh aligns much more accurately with the input image.
    }
    \label{fig_annotation_compare}
\end{figure*}

\noindent\textbf{Pose refinement.}
To ensure geometric consistency between camera poses and instance objects, we constrain ray sampling to instance mask regions. Following Co-SLAM~\cite{wang2023coslam}, we model camera rays as lines from origin $o_t$ with direction $r_{t,u,v}$, but sample only from pixels where $M_i(u,v)=1$:
\begin{equation}
\begin{aligned}
\mathcal{R}_t = &\left\{ (o_t, r_{t,u,v}) \mid \exists i: M_i(u,v)=1, \right. \\
&\left. (u,v) \in \{1,\ldots, W\} \times \{1,\ldots, H\} \right\}
\end{aligned}
\end{equation}
For each ray, we sample $M_c$ uniform points and $M_f$ depth-guided near-surface points within $[d_{\text{near}}, d_{\text{far}}]$. This mask-constrained sampling ensures that optimization signals derive solely from target instances, reducing background interference.

We jointly optimize camera pose $\xi_t$ and neural scene representation $f_\theta$ (mapping world coordinates to color and TSDF) via a multi-loss objective. Beyond the standard color, depth, SDF, and free-space losses from Co-SLAM, we introduce an instance mask alignment loss $\mathcal{L}_{\text{mask}}$ to enforce pose-instance geometric consistency:
\begin{equation}
\begin{aligned}
\mathcal{L}_{\text{mask}} = &\frac{1}{|\mathcal{R}_t|} \sum_{(o_t, r_{t,u,v}) \in \mathcal{R}_t} \left(1 - \max_i M_i(u, v)\right) \\
&\cdot \left\| \hat{d}_{t,u,v} - d_{t,u,v} \right\|_2^2
\end{aligned}
\end{equation}
where $\max_i M_i(u,v)=1$ for instance pixels (no penalty) and $0$ for background. The total loss combines all terms with weights $\lambda_{\text{rgb}}=5$, $\lambda_d=0.1$, $\lambda_{\text{sdf}}=1000$, $\lambda_{\text{fs}}=10$, and $\lambda_{\text{mask}}=2$:
\begin{equation}
\begin{aligned}
\mathcal{L}_{\text{total}} = &\lambda_{\text{rgb}} \mathcal{L}_{\text{rgb}} + \lambda_{d} \mathcal{L}_{d} + \lambda_{\text{sdf}} \mathcal{L}_{\text{sdf}} + \\
&\lambda_{\text{fs}} \mathcal{L}_{\text{fs}} + \lambda_{\text{mask}} \mathcal{L}_{\text{mask}}
\end{aligned}
\end{equation}

We perform global bundle adjustment to jointly optimize all camera poses $\{\xi_t\}_{t=1}^T$ and scene representation $f_\theta$ using rays sampled from all keyframe instance masks:
\begin{equation}
\begin{split}
\arg\min_{\theta, \{\xi_t\}} 
&\frac{1}{|\mathcal{R}_{\text{global}}|} \sum_{(o_t,r_{t,u,v}) \in \mathcal{R}_{\text{global}}} \\
&\mathcal{L}_{\text{total}}(\hat{c}_{t,u,v}, \hat{d}_{t,u,v}, M_i(u,v))
\end{split}
\end{equation}

Compared to the original Co-SLAM, we modify the strategy in two main aspects: (1) the keyframe ray pool is restricted to instance mask rays, thereby focusing the optimization on object geometry; and (2) the initial poses are obtained from FoundationPose instead of relying on constant-velocity assumptions, reducing the risk of convergence to local optima. We alternate between optimizing the scene parameters $\theta$ for $k = 10$ iterations and updating the poses using accumulated gradients. These modifications improve object-to-camera pose consistency.

\section{Experiments}
\label{sec:experiments}

In this section, we first present results of our 3D model alignment procedure in Section~\ref{models_alignment_results}. We then evaluate state-of-the-art 6D object pose estimation methods on DeSOPE across different deformation levels in Section~\ref{sec:sota_benchmark}. Finally, we analyze the factors contributing to performance degradation in Section~\ref{sec:factors_analysis}.

\subsection{Results of 3D Model Alignment}
\label{models_alignment_results}

We employ a multi-view matching strategy to refine the alignment between the deformed and undeformed 3D models following manual alignment. This process involves computing dense 2D–2D pixel correspondences between the two models from six orthogonal viewpoints, as described in Section~\ref{3D_col_align}. 

Figure~\ref{fig_align_results} presents qualitative results of the alignment. It demonstrates that the multi-view matching strategy effectively captures the deformation state of vertices.

Table~\ref{tab_average_align_results} reports quantitative results of the 3D model alignment. “Init.” denotes the matching error obtained from manual alignment, while “+Refine” indicates the error after refinement. The six intermediate rows correspond to the alignment errors from each of the six projected views, and the final row summarizes the overall alignment performance. The proposed automatic refinement significantly reduces the alignment error.

\begin{table}[t]
  \centering
    \begin{tabular}{cccccc}
    \toprule
     \multicolumn{2}{c}{Deformed 1}   & \multicolumn{2}{c}{Deformed 2} & \multicolumn{2}{c}{Deformed 3} \\
     Init. & +Refine & Init. & +Refine & Init. & +Refine \\
     \midrule
0.473  & {\bf 0.261} & 0.938  &  {\bf 0.440}  & 1.177  &  {\bf 1.010} \\
0.657  & {\bf 0.191}  & 1.050  &  {\bf 0.511}  & 1.306  &  {\bf 1.253} \\
1.922  & {\bf 1.129}  & 2.399  &  {\bf 1.358}  & 1.291  &  {\bf 1.091} \\
0.731  & {\bf 0.297} & 0.967  &  {\bf 0.577}  & 0.980  &  {\bf 0.843} \\
0.855  & {\bf 0.498}  & 1.098  &  {\bf 0.576}  & {\bf 0.672}  &  0.690 \\
0.988  & {\bf 0.651}  & 1.530  &  {\bf 0.931}  & 1.076  &  {\bf 0.878} \\
\midrule
0.782  & {\bf 0.538}  & 1.138  &  {\bf 0.719}  & 1.400  &  {\bf 0.933} \\
    \bottomrule
    \end{tabular}%
      \caption{{\bf Results of 3D model alignment.} ``Init.'' denotes the matching error obtained from manual alignment between deformed and undeformed objects, while ``+Refine'' indicates the error after refinement. The six middle rows report the alignment errors for each of the six projected views, and the final row summarizes the overall alignment error. The error is reported in centimeters after RANSAC outlier removal with a threshold of 3 cm.
      }
  \label{tab_average_align_results}%
\end{table}%

\subsection{Evaluation of State-of-the-Art Methods}
\label{sec:sota_benchmark}

We evaluate three RGB-D methods on our DeSOPE dataset: SCFlow2~\cite{wang2025scflow2}, FoundationPose~\cite{wen2024foundationpose}, and GenPose~\cite{genpose2023}. SCFlow2 and FoundationPose are designed for unseen object pose estimation without retraining at inference time, while GenPose is a typical category-level object pose method. Notably, SCFlow2 is the RGB-D extension of SCFlow~\cite{yang2023scflow} that is used in 3D model alignment.

For SCFlow2 and FoundationPose, we directly use the pre-trained models provided by the authors, which are trained on large-scale datasets. For GenPose, we retrain the model on our dataset by treating all four meshes (the canonical mesh and three deformed variants) as a single category, using the canonical mesh to represent the category shape.

All methods rely on a provided mesh during inference. In our setup, we use the same canonical mesh for all methods, regardless of whether the object in the image is deformed, which is consistent with the assumptions of these models. However, for metric computation, we use the actual mesh present in the image.

We adopt the standard evaluation metrics from the BOP challenge~\cite{hodan2018bop}, including Visible Surface Discrepancy (VSD), Maximum Symmetry-Aware Surface Distance (MSSD), and Maximum Symmetry-Aware Projection Distance (MSPD). VSD measures the agreement between estimated and ground-truth poses over visible surfaces, MSSD evaluates surface distance across the entire object, and MSPD quantifies projection error on the object surface. Following the BOP evaluation protocol, we report Average Recall (AR)—defined as the mean recall over VSD, MSSD, and MSPD—as the primary evaluation metric. We refer the reader to~\cite{hodan2018bop} for more details on the metrics.

\begin{figure}[t]
    \centering
    \setlength\tabcolsep{1pt}
    \begin{tabular}{ccc}
        \includegraphics[width=0.32\linewidth, height=0.36\linewidth]{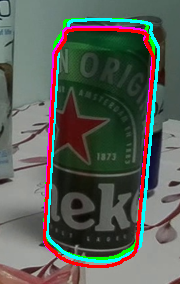} &
        \includegraphics[width=0.32\linewidth, height=0.36\linewidth]{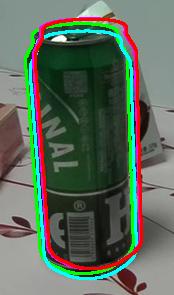} &
        \includegraphics[width=0.32\linewidth, height=0.36\linewidth]{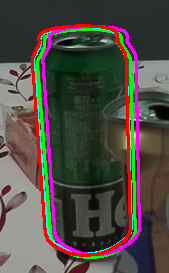} \\

        \includegraphics[width=0.32\linewidth, height=0.36\linewidth]{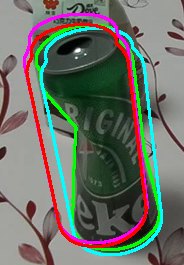} &
        \includegraphics[width=0.32\linewidth, height=0.36\linewidth]{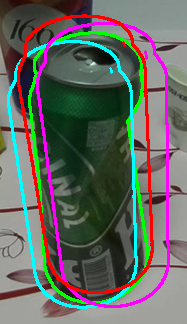} &
        \includegraphics[width=0.32\linewidth, height=0.36\linewidth]{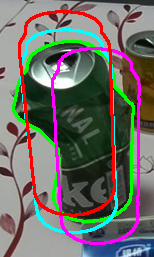} \\

    \end{tabular}
    \caption{{\bf State-of-the-art methods on DeSOPE.} Most methods achieve strong performance on images with canonical meshes (first row). However, their accuracy degrades significantly when the meshes undergo deformations that deviate from the canonical configuration (second row), as they assume the target in the image still conforms to the canonical mesh, which is not the case. Pose estimation results are projected onto the 2D plane using the corresponding mesh. Color code: green—Ground Truth; red—GenPose; pink—FoundationPose; cyan—SCFlow2.
    }
    \label{fig_sota_results}
\end{figure}

As shown in Table~\ref{tab_genpose} and Fig.~\ref{fig_sota_results}, all three methods perform well on images with the correct canonical mesh. However, their performance drops when applied to deformed objects, mainly because the models assume the mesh in the image is canonical, which is not the case under unknown deformations. We observe a similar trend in images with human manipulation. Overall, the more severe the deformation, the worse the performance in both settings.

\begin{table}[t]
  \centering

    \begin{tabular}{cccc}
    \toprule
    & SCFlow2 & FoundationPose & GenPose \\
    \midrule
Canonical & {\bf 0.82} &{\bf 0.78} &{\bf 0.67} \\
Deformed 1 & 0.67 &0.58 &0.56 \\
Deformed 2 & 0.43 &0.38 &0.36 \\
Deformed 3 & 0.23 &0.24 &0.31 \\
\midrule
Canonical & {\bf 0.77} &{\bf 0.72 }&{\bf 0.61} \\
Deformed 1 & 0.64 &0.54 &0.53 \\
Deformed 2 & 0.34 &0.30 &0.37 \\
Deformed 3 & 0.20 &0.20 &0.28 \\
    \bottomrule
    \end{tabular}%
      \caption{{\bf Evaluation of state-of-the-art methods on DeSOPE.}
      We report results in two groups: the first is evaluated on all the images without human manipulation, and the second on images with human manipulation.
      We report the Average Recall (AR) for three representative methods: SCFlow2~\cite{wang2025scflow2}, FoundationPose~\cite{wen2024foundationpose}, and GenPose~\cite{genpose2023}. Although these methods generalize well, their performance degrades significantly when the meshes in the images do not match the canonical mesh assumed by the model. The more severe the deformation, the worse their performance.
      }
  \label{tab_genpose}%
\end{table}%

\subsection{Performance Analysis}
\label{sec:factors_analysis}
\begin{figure}[t]
    \centering
    \begin{subfigure}[t]{0.49\textwidth}
        \centering
        \begin{tikzpicture}[scale=0.54]
        \begin{axis}[
            ymin=0, ymax=1.0,
            xmin=-0.5, xmax=3.5,
            ytick={0,0.2,0.4,0.6,0.8,1.0},
            xtick={0,1,2,3},
            xticklabels={Canonical, Deformed 1, Deformed 2, Deformed 3},
            legend pos=south west,
            grid=both,
            grid style={dashed, gray!30},
            label style={font=\large},
            legend style={font=\large, legend cell align=left}
        ]
        \addplot[color=blue, mark=o, line width=1pt] coordinates {
            (0, 0.94) (1, 0.83) (2, 0.56) (3, 0.33)
        };
        \addlegendentry{SCFlow2}
        \addplot[color=green!70!black, mark=square, line width=1pt] coordinates {
            (0, 0.90) (1, 0.71) (2, 0.47) (3, 0.31)
        };
        \addlegendentry{FPose}
        \addplot[color=red, mark=triangle, line width=1pt] coordinates {
            (0, 0.84) (1, 0.69) (2, 0.51) (3, 0.39)
        };
        \addlegendentry{GenPose}
        \end{axis}
        \end{tikzpicture}
        \hspace{0.1cm}
        \begin{tikzpicture}[scale=0.54]
        \begin{axis}[
            ymin=0, ymax=1.0,
            xmin=-0.5, xmax=3.5,
            ytick={0,0.2,0.4,0.6,0.8,1.0},
            xtick={0,1,2,3},
            xticklabels={Canonical, Deformed 1, Deformed 2, Deformed 3},
            legend pos=south west,
            grid=both,
            grid style={dashed, gray!30},
            label style={font=\large},
            legend style={font=\large, legend cell align=left}
        ]
        \addplot[color=blue, mark=o, line width=1pt] coordinates {
            (0, 0.79) (1, 0.64) (2, 0.37) (3, 0.18)
        };
        \addlegendentry{SCFlow2}
        \addplot[color=green!70!black, mark=square, line width=1pt] coordinates {
            (0, 0.74) (1, 0.55) (2, 0.32) (3, 0.19)
        };
        \addlegendentry{FPose}
        \addplot[color=red, mark=triangle, line width=1pt] coordinates {
            (0, 0.64) (1, 0.55) (2, 0.36) (3, 0.29)
        };
        \addlegendentry{GenPose}
        \end{axis}
        \end{tikzpicture}
        \caption{Under no occlusion and 50\% occlusion}
    \end{subfigure}

    \vspace{0.3em}

    \begin{subfigure}[t]{0.49\textwidth}
        \centering
        \begin{tikzpicture}[scale=0.54]
        \begin{axis}[
            ymin=0, ymax=1.0,
            xmin=-0.5, xmax=3.5,
            ytick={0,0.2,0.4,0.6,0.8,1.0},
            xtick={0,1,2,3},
            xticklabels={Canonical, Deformed 1, Deformed 2, Deformed 3},
            legend pos=south west,
            grid=both,
            grid style={dashed, gray!30},
            label style={font=\large},
            legend style={font=\large, legend cell align=left}
        ]
        \addplot[color=blue, mark=o, line width=1pt] coordinates {
            (0, 0.90) (1, 0.73) (2, 0.52) (3, 0.27)
        };
        \addlegendentry{SCFlow2}
        \addplot[color=green!70!black, mark=square, line width=1pt] coordinates {
            (0, 0.83) (1, 0.72) (2, 0.40) (3, 0.25)
        };
        \addlegendentry{FPose}
        \addplot[color=red, mark=triangle, line width=1pt] coordinates {
            (0, 0.74) (1, 0.65) (2, 0.45) (3, 0.35)
        };
        \addlegendentry{GenPose}
        \end{axis}
        \end{tikzpicture}
        \hspace{0.1cm}
        \begin{tikzpicture}[scale=0.54]
        \begin{axis}[
            ymin=0, ymax=1.0,
            xmin=-0.5, xmax=3.5,
            ytick={0,0.2,0.4,0.6,0.8,1.0},
            xtick={0,1,2,3},
            xticklabels={Canonical, Deformed 1, Deformed 2, Deformed 3},
            legend pos=south west,
            grid=both,
            grid style={dashed, gray!30},
            label style={font=\large},
            legend style={font=\large, legend cell align=left}
        ]
        \addplot[color=blue, mark=o, line width=1pt] coordinates {
            (0, 0.83) (1, 0.71) (2, 0.45) (3, 0.20)
        };
        \addlegendentry{SCFlow2}
        \addplot[color=green!70!black, mark=square, line width=1pt] coordinates {
            (0, 0.78) (1, 0.55) (2, 0.36) (3, 0.24)
        };
        \addlegendentry{FPose}
        \addplot[color=red, mark=triangle, line width=1pt] coordinates {
            (0, 0.70) (1, 0.60) (2, 0.41) (3, 0.33)
        };
        \addlegendentry{GenPose}
        \end{axis}
        \end{tikzpicture}
        \caption{Without and with human manipulation (under 20\% occlusion)}
    \end{subfigure}
    \caption{
    {\bf Performance analysis.} All plots present Average Recall (AR) across four mesh sets (Canonical and Deformed 1–3) for three methods: SCFlow2, FoundationPose (FPose), and GenPose. Key observations: (1) performance decreases as deformation severity increases across all settings; (2) greater occlusion leads to lower performance; (3) scenes with human manipulation consistently perform worse due to complex motion and occlusions during interaction.
    }
    \label{fig_factor_occlusion}
\end{figure}
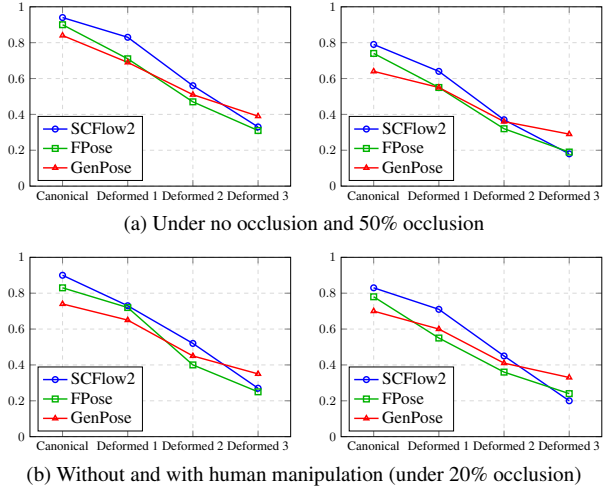

In this section, we evaluate model performance under additional factors. We first group the dataset by different occlusion ratios and report the performance of baseline methods. As shown in Fig.~\ref{fig_factor_occlusion}(a), performance drops significantly as deformation increases, and higher occlusion further degrades accuracy.

We also evaluate the effect of human manipulation under a setting of mild occlusion (20\%). As illustrated in Fig.~\ref{fig_factor_occlusion}(b), all three methods consistently achieve lower accuracy in scenes with human manipulation across all deformation levels. This performance gap can be attributed to two main factors: (1) hand occlusions during manipulation reduce the visible surface area, limiting the geometric cues available for pose estimation; and (2) motion blur caused by rapid hand movements degrades the quality of RGB-D observations, affecting both feature extraction and depth estimation. Notably, performance degradation under human manipulation is more pronounced for methods that rely heavily on precise geometric matching, such as SCFlow2 and FoundationPose, whereas GenPose exhibits relatively smaller degradation due to the retraining on the dataset and its category-level generalization capability.

In general, model performance is influenced by multiple factors. Across all the conditions we evaluated, the trend is consistent: the more severe the deformation, the worse the performance.

\section{Conclusion}

We introduced DeSOPE, the first dataset designed to study 6D object pose estimation under deformation, addressing a critical gap in existing benchmarks. Through comprehensive evaluation, we demonstrate that current state-of-the-art methods suffer performance degradation as deformation increases, revealing a fundamental limitation of rigid-object assumptions. By providing aligned 3D meshes across deformation states and large-scale pose annotations, DeSOPE establishes a challenging and realistic benchmark. We hope this dataset will encourage future research on deformation-aware representations, temporal modeling, and more robust pose estimation methods for real-world applications.

\vspace{0.5em}

\noindent {\bf Acknowledgments.}
This work was supported in part by the National Natural Science Foundation of China under Grant No. 62371359; the Youth Innovation Team of Shaanxi Universities; the ``Scientist + Engineer'' Team of Qin Chuang Yuan, Shaanxi Province; the Xi’an Science and Technology Program; the ``Leading the Charge'' Initiative for the Industrialization of Core Technologies in Key Industrial Chains in Shaanxi Province; the Key Research and Development Program of Shaanxi (Grant No. 2024GX-ZDCYL-02-09); and the Key Research Program of the Chinese Academy of Sciences (Grant No. KGFZD-145-2023-15).

{
    \small
    \bibliographystyle{ieeenat_fullname}
    \bibliography{main}
}

\end{document}